\documentclass[sigconf,nonacm]{acmart}

\usepackage{epsfig}
\usepackage{graphicx}
\usepackage{amsmath}
\usepackage{verbatim} 
\usepackage{changepage}
\usepackage{float}

\usepackage{relsize}
\usepackage[T1]{fontenc} 

\usepackage{tikz}
\usetikzlibrary{shapes.geometric, arrows.meta, positioning, fit, backgrounds, shadows, decorations.pathmorphing}
\usepackage{caption}
\usepackage{midfloat}
\usepackage{enumitem}
\usepackage{fancyvrb}
\newcommand\userinput[1]{\textbf{#1}}
\usepackage{tgcursor}
\usepackage{booktabs} 
\usepackage{array}    

\usepackage{color}
\begin{document}

\title{SynLexLM: Scaling Legal LLMs with Synthetic Data and Curriculum Learning}

\author{Ojasw Upadhyay}
\orcid{0000-0003-4779-3063}
\affiliation{%
  \institution{Georgia Institue of Technology}
  \city{Atlanta}
  \state{Georgia}
  \country{USA}
}
\email{ojasw@gatech.com}

\author{Abishek Saravanakumar}
\affiliation{%
  \institution{Georgia Institue of Technology}
  \city{Atlanta}
  \state{Georgia}
  \country{USA}
}
\email{abishek@gatech.edu}

\author{Ayman Ismail}
\affiliation{%
  \institution{Georgia Institue of Technology}
  \city{Atlanta}
  \state{Georgia}
  \country{USA}
}
\email{ayman@gatech.edu}

\renewcommand{\shortauthors}{Upadhyay et al.}
\acmConference[arXiv '25]{}{April 26,
  2025}{Atlanta, GA}

\begin{abstract}
   \textit{Large Language Models (LLMs) \cite{AttentionIsAllYouNeed} are powerful but often require extensive fine-tuning and large datasets for specialized domains like law. General-purpose pre-training may not capture legal nuances, and acquiring sufficient legal data is challenging \cite{SynthDataTextbook}. We introduce SynLexLM, a novel approach to efficiently pre-train a legal LLM. Our method employs curriculum learning \cite{CurriculumLearning}, progressing from simple to complex legal texts and queries, combined with synthetic data augmentation \cite{SynthDataTextbook, Guan2018} using models like Gemini Pro to address data scarcity. We aim to achieve improved performance on legal benchmarks (BigLaw-Bench \cite{HarveyBigLaw}, EUR-Lex-Sum \cite{EurLex-Sum}) compared to traditional models and fine-tuned versions \cite{Finetuning}. Preliminary work involves generating synthetic QA pairs reflecting legal reasoning. This work aims to enhance legal document analysis and research tools, potentially democratizing access to advanced legal AI.}
\end{abstract}

\begin{abstract}
   \textit{Large Language Models (LLMs) \cite{AttentionIsAllYouNeed} are powerful but often require extensive fine-tuning and large datasets for specialized domains like law. General-purpose pre-training may not capture legal nuances, and acquiring sufficient legal data is challenging \cite{SynthDataTextbook}. We introduce SynLexLM, a novel approach to efficiently pre-train a legal LLM. Our method employs curriculum learning \cite{CurriculumLearning}, progressing from simple to complex legal texts and queries, combined with synthetic data augmentation \cite{SynthDataTextbook, Guan2018} using models like Gemini Pro to address data scarcity. We aim to achieve improved performance on legal benchmarks (BigLaw-Bench \cite{HarveyBigLaw}, EUR-Lex-Sum \cite{EurLex-Sum}) compared to traditional models and fine-tuned versions \cite{Finetuning}. Preliminary work involves generating synthetic QA pairs reflecting legal reasoning. This work aims to enhance legal document analysis and research tools, potentially democratizing access to advanced legal AI.}
\end{abstract}

\maketitle


\section*{Introduction}

Large Language Models (LLMs) \cite{AttentionIsAllYouNeed} have revolutionized natural language processing, yet their direct application to highly specialized fields like law remains challenging. Legal language is characterized by its precision, reliance on specific terminology, complex reasoning structures, and dependence on vast amounts of contextual information (statutes, case law). General-purpose LLMs, typically pre-trained on broad web corpora, often fail to capture these nuances effectively \cite{SynthDataTextbook}. Furthermore, fine-tuning these models for legal tasks \cite{Finetuning} is hampered by the limited availability of large-scale, high-quality, annotated legal datasets, often due to privacy concerns, cost, or the specialized nature of the data.

This project tackles the core problem: \textbf{How can we effectively train or adapt LLMs for the legal domain despite the scarcity of real-world legal data?} Our proposed solution is \textbf{SynLexLM}, an approach centered around two key strategies:
\begin{enumerate}
    \item \textbf{Synthetic Data Augmentation:} We leverage advanced generative models, Google's Gemini Pro, to create synthetic legal text, focusing on question-answer pairs derived from real legal documents. This artificially expands the training data pool, targeting specific legal reasoning skills \cite{Guan2018, SynthDataTextbook}.
    \item \textbf{Curriculum Learning:} Inspired by human learning, we structure the training process by presenting the model with data in increasing order of complexity \cite{CurriculumLearning}. We start with simpler legal texts and queries (based on factors like document length, concept density, and question type) and gradually introduce more difficult ones.
\end{enumerate}
Our specific goal is to demonstrate that this combined approach allows us to train a Gemma 3 12b that achieves competitive or improved performance on established legal NLP benchmarks, such as BigLaw-Bench \cite{HarveyBigLaw} and EUR-Lex-Sum \cite{EurLex-Sum}.

The development of effective and data-efficient legal LLMs has significant implications. Success would mean:
\begin{itemize}
    \item \textbf{Enhanced Legal Practice Efficiency:} Tools built on such models could drastically reduce the time spent by legal professionals on tasks like document review, summarization, research, and due diligence, potentially lowering costs and improving service delivery.
    \item \textbf{Democratization of Legal AI:} By reducing the reliance on massive proprietary datasets, our approach could make sophisticated legal AI tools more accessible to smaller firms, public defenders, legal aid organizations, and even individuals navigating the legal system.
    \item \textbf{Improved Consistency and Accuracy:} AI tools can assist in maintaining consistency in legal analysis and potentially reduce errors in complex document handling.
    \item \textbf{Advancement in Domain-Specific AI:} Showing success in the challenging legal domain provides insights and methods applicable to other specialized, data-scarce domains like medicine or finance.
\end{itemize}
Ultimately, SynLexLM aims to contribute to a future where advanced AI meaningfully assists in legal understanding and practice, while also necessitating careful consideration of ethical implications like bias, fairness, and transparency. 

\section*{Related Works}

Our research integrates concepts from synthetic data generation, curriculum learning, large language models, and legal NLP benchmarks.

\textbf{Synthetic Data Generation:} The challenge of data scarcity in specialized domains has spurred research into synthetic data. Nikolenko's work \cite{SynthDataTextbook} provides a foundational overview of techniques for generating synthetic data for deep learning, particularly relevant in domains like law where data can be scarce or sensitive. While various methods exist, generative models are increasingly used. Guan et al. \cite{Guan2018} showed the potential of generating synthetic text in the medical domain. Although sequence GANs combined with reinforcement learning (e.g., SeqGAN \cite{SeqGAN}) offer an alternative for text generation, their computational demands led us to prefer leveraging large pre-trained models like Gemini Pro  for controlled, high-quality generation via prompting.

\textbf{Curriculum Learning:} The idea of training models on progressively harder examples, mimicking human learning, was formalized by Bengio et al. \cite{CurriculumLearning}. Curriculum learning has been shown to improve convergence speed and generalization performance in various tasks. We apply this by ordering our combined real and synthetic legal data based on complexity metrics like document length and legal concept density.

\textbf{Large Language Models and Domain Adaptation:} Our work inherently relies on the power of Transformer-based LLMs \cite{AttentionIsAllYouNeed}, such as Llama \cite{Llama3.2} and Gemma 3 12b. Adapting these generalist models to specific domains is a key research area. Techniques like ULMFiT \cite{Finetuning} demonstrated effective fine-tuning strategies for adapting pre-trained language models to downstream tasks. Our approach extends fine-tuning by incorporating both targeted synthetic data generation and a curriculum strategy specifically for the complexities of the legal domain.

\textbf{Legal NLP Benchmarks:} To evaluate our approach rigorously, we rely on established benchmarks designed for legal language understanding. EUR-Lex-Sum \cite{EurLex-Sum} focuses on long-form summarization in the EU legal context. Broader benchmarks like LexGLUE \cite{LexGLUE} and the industry-focused BigLaw-Bench \cite{HarveyBigLaw} provide standardized tasks for evaluating capabilities such as question answering, text classification, and named entity recognition within the legal field.

\section*{Method / Approach}

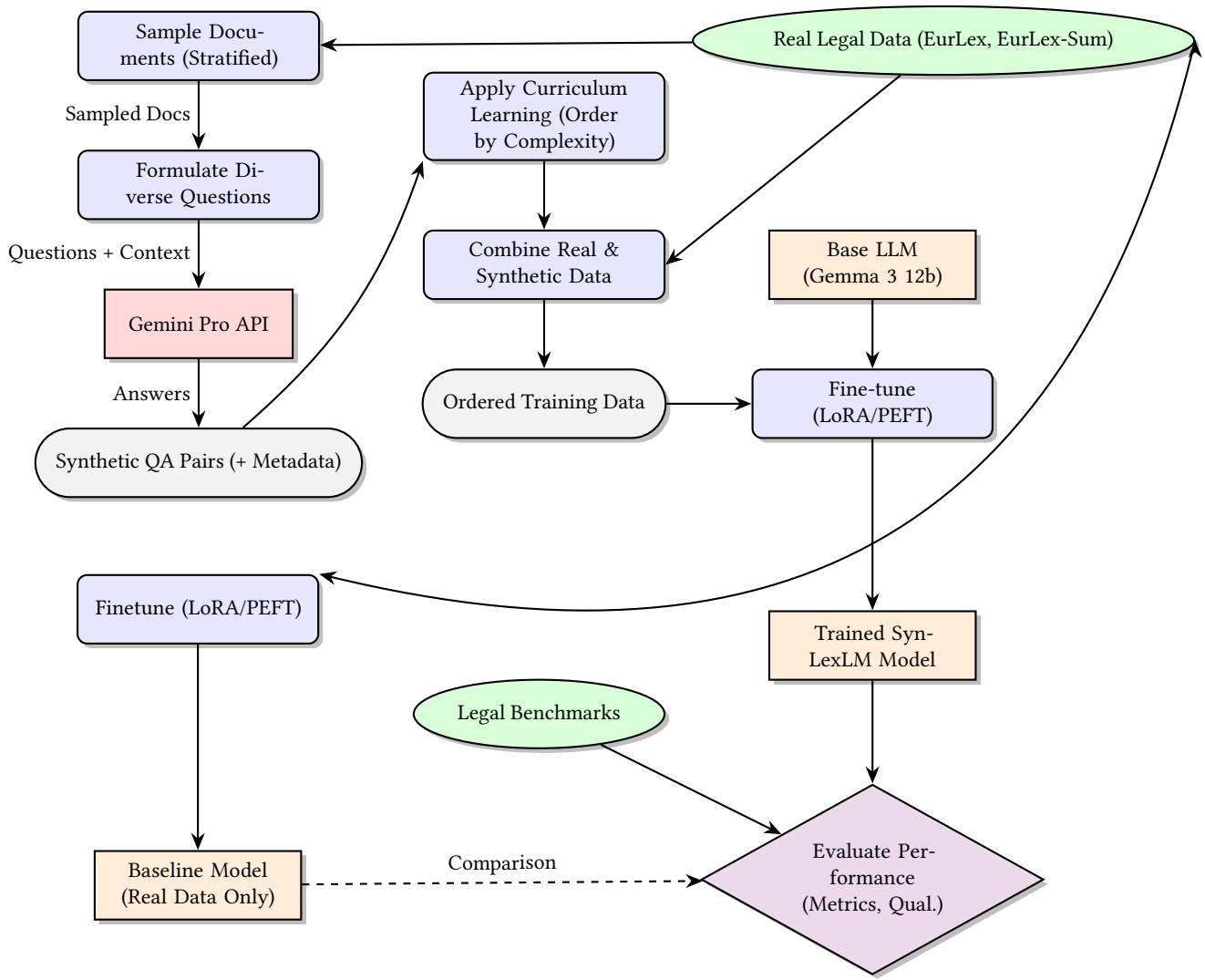
\begin{figure*}[!htbp] 
\centering
\begin{tikzpicture}[
    node distance=10mm and 15mm, 
    base/.style = {draw, thick, minimum height=10mm, text centered, drop shadow}, 
    process/.style = {base, rectangle, rounded corners, minimum width=35mm, text width=33mm, fill=blue!10},
    data/.style = {base, ellipse, fill=green!15},
    model/.style = {base, rectangle, fill=orange!15, minimum width=30mm, text width=28mm},
    tool/.style = {base, fill=red!15, minimum width=1mm},
    eval/.style = {base, diamond, aspect=1.8, fill=violet!15, text width=25mm},
    io/.style = {base, rectangle, rounded corners=5mm, inner sep=3mm, fill=gray!10},
    arrow/.style = {-{Stealth[length=2.5mm, width=2mm]}, thick},
    dashedarrow/.style = {arrow, dashed}
]

\node (real_data) [data] {Real Legal Data (EurLex, EurLex-Sum)};
\node (sampling) [process, below left=of real_data, yshift=18mm, xshift=-50mm] {Sample Documents (Stratified)};
\node (question_gen) [process, below=of sampling] {Formulate Diverse Questions};
\node (gemini) [tool, below=of question_gen,inner sep=10pt] {Gemini Pro API};
\node (synth_qa) [io, below=of gemini] {Synthetic QA Pairs (+ Metadata)};

\node (curriculum) [process, right=of question_gen, yshift=10mm] {Apply Curriculum Learning (Order by Complexity)};
\node (combine) [process, below=of curriculum, xshift=0mm] {Combine Real \& Synthetic Data};
\node (train_data) [io, below=of combine] {Ordered Training Data};

\node (base_llm) [model, right=of combine, xshift=0mm] {Base LLM (Gemma 3 12b)};
\node (lora) [process, below=of base_llm, text width=25mm] {Fine-tune (LoRA/PEFT)};

\node (synlexlm) [model, below=of lora, yshift=-15mm] {Trained SynLexLM Model};

\node (baseline) [model, left=of synlexlm, xshift=-53mm, yshift=-35mm] {Baseline Model (Real Data Only)};
\node (base_lora) [process, above=of baseline, yshift=20mm] {Finetune (LoRA/PEFT)};
\node (benchmarks) [data, left=of synlexlm, xshift=0mm, yshift=-10mm] {Legal Benchmarks};
\node (evaluation) [eval, below=of synlexlm, yshift=-5mm] {Evaluate Performance (Metrics, Qual.)};

\draw [arrow] (real_data) -- (sampling);
\draw [arrow] (real_data) -- (combine.east);
\draw [arrow] (sampling) -- node[midway, left, align=center] {Sampled Docs} (question_gen);
\draw [arrow] (question_gen) -- node[midway, left] {Questions + Context} (gemini);
\draw [arrow] (gemini) -- node[midway, left] {Answers} (synth_qa);
\draw [arrow] (synth_qa) edge[bend right=15] (curriculum.south west);
\draw [arrow] (curriculum) -- (combine);
\draw [arrow] (combine) -- (train_data);
\draw [arrow] (train_data) -- (lora);
\draw [arrow] (real_data.east) edge[bend left=45, looseness=1.23] (base_lora.north east);
\draw [arrow] (base_lora) -- (baseline);
\draw [arrow] (base_llm) -- (lora);
\draw [arrow] (lora) -- (synlexlm);
\draw [arrow] (synlexlm) -- (evaluation);
\draw [dashedarrow] (baseline) -- node[midway, above, sloped] {Comparison} (evaluation);
\draw [arrow] (benchmarks) -- (evaluation);


\end{tikzpicture}
\caption{Architecture diagram of the SynLexLM system, illustrating the flow from real data curation through synthetic data generation, curriculum learning, model fine-tuning, and evaluation.}
\label{fig:synlexlm_architecture}
\end{figure*}

SynLexLM employs a multi-faceted approach to train a legal LLM effectively with limited real data. The core components are data curation, synthetic data generation via prompting, and curriculum-based training as shown in Figure \ref{fig:synlexlm_architecture}.

\subsection*{Data Curation}
We curated a foundational dataset of real legal text from established sources:
\begin{itemize}
    \item \textbf{EurLex}: Containing over 55,000 labeled entries \cite{LexGLUE}.
    \item \textbf{EurLex-Sum}: Over 11,000 documents paired with summaries \cite{EurLex-Sum}.
\end{itemize}
This real data serves as the basis for training and as seed material for synthetic generation.

\subsection*{Synthetic Data Generation via Gemini Pro}
To augment the real data, we generate synthetic Question-Answer (QA) pairs using Google's Gemini 2.5 Pro API:

\textbf{Process:} We sample documents from our curated real dataset. Each sampled document is processed by a multi-strategy module designed to formulate diverse questions targeting:
    \begin{itemize}
        \item Factual Extraction
        \item Definitions
        \item Legal Reasoning
        \item Comparisons
    \end{itemize}

\textbf{Prompting:} The generated question, along with relevant context from the source document, is used to prompt Gemini Pro to generate an answer. API interactions are managed to handle rate limits at 15 requests/min.

\textbf{Augmentation:} Each generated QA pair is tagged with metadata from the source document and an estimated difficulty level based on the generation strategy (e.g., factual vs. reasoning).
This allows us to create targeted training examples that exercise specific legal reasoning skills.

\subsection*{Curriculum Learning Framework}
We structure the training data (both real and synthetic) using a curriculum learning approach \cite{CurriculumLearning}:
\begin{itemize}
    \item \textbf{Complexity Metrics:} Data points are ordered based on complexity, primarily considering document length and estimated legal concept density.
    \item \textbf{Stratification:} To ensure diversity and prevent skewness, we employ stratified sampling when selecting documents for synthetic generation and potentially during batch creation. Stratification considers factors like document length (categorized as short, medium, long) to maintain proportions similar to the real dataset.
    \item \textbf{Staged Training:} The model is exposed to data in increasing order of difficulty, starting with simpler examples and progressing to more complex ones.
\end{itemize}

\subsection*{Model and Training}
Our approach is designed to be applied to transformer-based LLMs. Our pipeline utilizes Gemma 3 12b and combines curated real data and generated synthetic data with a pre-training and fine-tuning mix.

\textbf{Justification and Alternatives:}
This combined approach is motivated by the need to overcome data scarcity (addressed by synthetic data) while ensuring effective learning of complex legal concepts (aided by curriculum learning). Using a powerful API like Gemini Pro for generation is more resource-efficient for this project than training a custom generator (like a GAN+RL model \cite{SeqGAN}, which was considered but deemed computationally prohibitive). The alternative of simple fine-tuning on only real data serves as our primary baseline but is expected to be limited by data availability.


\textbf{Use of Existing Resources:}
We explicitly utilize the Google Gemini 2.5 Pro API, Gemma 3, and publicly available datasets (EurLex \cite{EurLex-Sum}, EurLex-Sum \cite{EurLex-Sum}). Our work builds upon existing LLM architectures. The novelty lies in the specific combination and application of synthetic data generation and curriculum learning strategies tailored to the legal domain. 

\section*{Data and Pre-Processing}

Our project utilizes a combination of real legal datasets and synthetically generated data derived from them.

\subsection*{Real Datasets}
The primary sources of authentic legal text are:
\begin{itemize}
    \item \textbf{EurLex}: A dataset that comprises of over 55,000 legal documents from the European Union, each associated with specific labels (metadata). Sourced from official EU publications. \cite{EurLex-Sum}
    \item \textbf{EurLex-Sum}: Contains over 11,000 documents from EurLex, paired with human-written summaries and Celex IDs. This dataset is crucial for training for summarization tasks. \cite{EurLex-Sum}
    \item \textbf{Other Potential Sources (for fine-tuning):} As outlined in future work (Section \ref{sec:conclusion}), there is potential to leverage additional datasets for broader fine-tuning, potentially including German legal NER \cite{GermanNER}, other datasets within LexGLUE \cite{LexGLUE}, US legal texts (i.e. SCOTUS), and an Australian legal classification dataset \cite{KaggleLegal}.
\end{itemize}

These datasets provide the grounding in real legal language and structure.

\subsection*{Synthetic Dataset}
The synthetic dataset is generated to supplement the "real" datasets, particularly targeting QA and summarization capabilities with the following details:
\begin{itemize}
    \item \textbf{Source:} Derived from documents sampled from EurLex and EurLex-Sum.
    \item \textbf{Content:} Consists mainly of Question-Answer pairs related to the source documents.
    \item \textbf{Size:} Scaled to significantly augment the real data. Sampled for 100 texts per dataset due to API and compute per time restraints.
    \item \textbf{Properties \& Preprocessing:} Each QA pair includes source metadata and an estimated difficulty level. The generation pipeline involved loading datasets, extracting text, formatting for API input, stratified sampling (based on document length: short, medium, long) to ensure diversity matching the real data distribution, generating QA pairs, and storing them. No complex text normalization beyond standard LLM tokenization was applied. 
    \item \textbf{Known Limitations:} Quality depends on the generator model (Gemini 2.5 Pro) and prompting. Potential for replicating biases from source data. Factual/legal accuracy requires ongoing validation.
\end{itemize}

\textbf{Data Splits:}
For training and evaluation, we intend to use standard practices. For benchmarks like LexGLUE or datasets like EurLex-Sum that have predefined splits, we will adhere to them. For other scenarios, we will use typical train/validation/test splits (e.g., 80\% / 10\% / 10\%) created from the combined real and synthetic data, ensuring no overlap between splits, especially regarding source documents for synthetic data.

\section*{Experiments and Results}

This section outlines the experiments conducted or planned to evaluate SynLexLM and presents preliminary findings. Our goal is to demonstrate that synthetic data combined with curriculum learning improves performance on legal tasks compared to baselines.

\subsection*{Experimental Setup}
\textbf{Baselines:}
    \begin{enumerate}
        \item \textit{Real Data Only (No CL/Synth):} A model fine-tuned solely on the curated real legal data (EurLex, EurLex-Sum). An initial experiment involved training Gemma 3 12b on sampled EurLex data for labeling.
        \item \textit{Traditional Fine-tuning (No CL/Synth):} Comparing against standard fine-tuning approaches referenced in literature \cite{Finetuning} if applicable benchmark results exist.
    \end{enumerate}
\textbf{Evaluation Metrics:}
    \begin{itemize}
        \item \textit{Quantitative:} Model performance on summarization tasks (on EurLex-Sum \cite{EurLex-Sum}) and classification and Q\&A tasks (on LexGLUE \cite{LexGLUE}, BigLaw-Bench \cite{HarveyBigLaw}). We also plan to assess efficiency (training time, data usage).
        \item \textit{Qualitative:} Manual inspection of generated outputs (summaries, answers) for relevance, accuracy, coherence, and legal sensibility. Identifying success/failure modes.
    \end{itemize}
\textbf{Evaluation Datasets:} Performance will be measured on tasks from EUR-Lex-Sum \cite{EurLex-Sum}, LexGLUE \cite{LexGLUE}, BigLaw-Bench \cite{HarveyBigLaw}, and potentially CUAD \cite{CUAD}.

\subsection*{Implementation Details}


\begin{itemize}
    \item \textit{Models:} Our fine-tuning methodology utilized \textbf{unsloth/gemma-3-12b-it} as the base LLM. Using Gemini Pro, we were able to generate question-answer pairs.
    \item \textit{Framework:} The fine-tuning process was implemented using the Unsloth framework, along with PyTorch and Hugging Face's TRL (Transformer Reinforcement Learning) library. We also utilized LoRA (Low-Rank Adaptation) \cite{LoRA} for parameter-efficient fine-tuning (PEFT), with the configuration \verb|r=8| (rank), \verb|lora\_alpha=16| (scaling factor), and \verb|lora\_dropout=0| (dropout rate). The finetuning targeted the q\_proj, k\_proj, v\_proj, o\_proj, gate\_proj, up\_proj, and down\_proj modules. The model was loaded in 8-bit precision (load\_in\_8bit=True) to optimize memory usage.
    \item \textit{Hyperparameters:} Some of the key parameters that we adjusted were setting the maximum sequence length to 8192 (16384 for EurLex-Sum due to large documents), using a batch size of 8, utilizing the AdamW 8-bit optimizer, setting the learning rate of 2e-5 with a linear scheduler and 5 warmup steps, and weight decay to 0.01. 
    \item \textit{Hardware:} Nvidia H100/H200 through PACE/ICE.
\end{itemize}

\subsection*{Preliminary Results and Analysis}
Current results focus on the successful setup and initial outputs of the data generation process. Rigorous quantitative evaluation is ongoing.

\begin{itemize}
    \item \textbf{Synthetic QA Generation:} The pipeline using Gemini Pro successfully generated a diverse set of 930 QA pairs based on different strategies (factual, definitional, etc.) from EurLex and EurLex-Sum documents. Stratification based on document length was implemented to ensure questions were generated from a variety of document lengths.
    \item \textbf{Qualitative Examples:} The generated QA pairs (examples in Figure \ref{fig:qa_examples_qualitative}) were relevant to the source text and cover different aspects of legal understanding, as shown in the examples below. This provides initial evidence that the synthetic data would be beneficial for training.

    \begin{figure}[ht]
    \begin{adjustwidth}{-1cm}{}
    \begin{Verbatim}[commandchars=\\\{\}]
      \userinput{Question:} When did the event described
      in the document occur?
      \userinput{Answer:} "The event described in the
      document, the approval of the outline
      plan of agricultural advisory work in
      Italy, occurred on 23 March 1981."

      \userinput{Question:} What comparison or contrast is
      being made in this section of the
      document? [Source: Commission Reg...]
      \userinput{Answer:} This section of the document
      primarily contrasts the current
      Commission Regulation (EC) No 1241/2001
      with a previous one, Commission
      Regulation (EC) No 3223/94. Here's how
      the comparison/contrast is made:
      **Relationship:** Regulation (EC) No
      1241/2001 is based on and refers back
      to Regulation (EC) No 3223/94. The
      current regulation establishes "standard
      import values" based on...
    \end{Verbatim}
    \end{adjustwidth}
    \caption{Examples of synthetic QA pairs generated from legal documents using Gemini Pro. Some parts of the question and responses are truncated for brevity with ellipses.}
    \label{fig:qa_examples_qualitative}
    \end{figure}

    \item \textbf{Baseline Setup:} The baseline experiment using Gemma 3-12b on real EurLex and EurLex-Sum data was configured, providing a reference point for future comparison.
    \item \textbf{Quantitative Evaluation:} Final quantitative results comparing SynLexLM against the baselines on the target benchmarks are pending the completion of the fine-tuning and evaluation runs planned on other datasets (see Conclusion). Table \ref{tab:results_placeholder} shows the results from EurLex and EurLex-Sum finetuning runs over 10 epochs.
\end{itemize}

\begin{table}[ht]
\begin{center}
\begin{tabular}{|l|c|c|}
\hline
 &  EurLex & EurLex-Sum \\ 
\hline\hline
Baseline & 0.1918 & 0.1639 \\
SynLexLM & \textbf{0.0152} & \textbf{0.0026} \\
\hline
\end{tabular}
\vspace*{0.25cm}
\caption{Comparison of final training losses after 10 epochs between Baseline (only data from datasets) and SynLexLM (combination of synthetic data generation and curriculum learning).}
\label{tab:results_placeholder}
\end{center}
\end{table}


\subsection*{Discussion}
The project has successfully demonstrated the feasibility of the proposed synthetic data generation pipeline for the legal domain. Qualitative results are promising. However, the ultimate success of SynLexLM relies on demonstrating statistically significant quantitative improvements over baselines on rigorous benchmarks. Challenges include ensuring the legal and factual correctness of synthetic data at scale and effectively tuning the curriculum learning parameters. The next steps focus on executing the planned evaluations to provide this evidence. \\

\begin{figure}[H]
\centering
  \includegraphics[width=1\columnwidth]{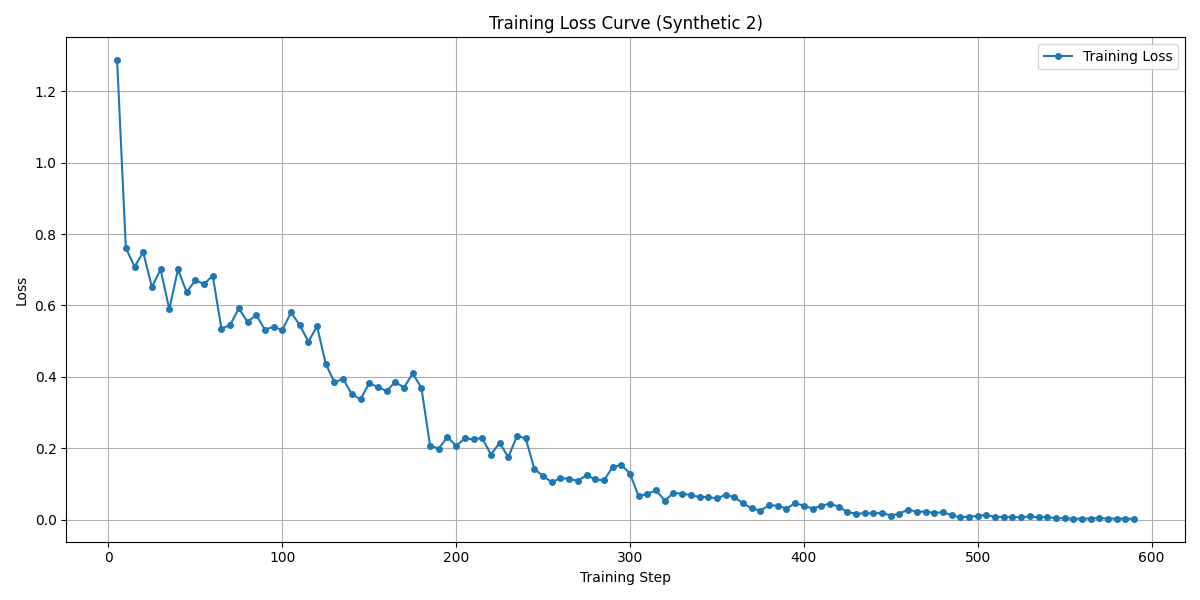}
  \caption{EurLex-Sum Training Losses Plot}
  \label{fig:EurLex plot}
\end{figure}

\begin{figure}[H]
\centering
  \includegraphics[width=1\columnwidth]{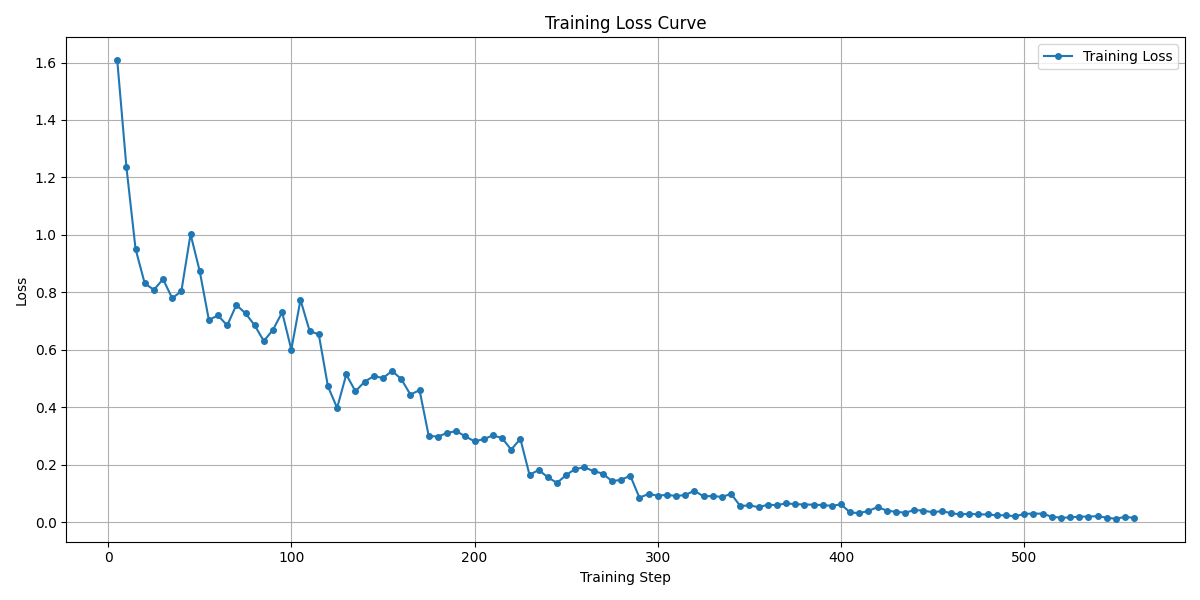}
  \caption{EurLex Training Losses Plot}
  \label{fig:EurLex-Sum plot}
\end{figure}

The training loss curves for the EurLex-Sum and EurLex datasets are in Figures \ref{fig:EurLex plot} and \ref{fig:EurLex-Sum plot}. The final loss we obtained on EurLex and EurLex-Sum with SynLexLM is 0.0152 and 0.0026 respectively. This is a significant improvement on the Baseline model that achieved a loss of 0.1639 and 0.0152 on EurLex and EurLex-Sum respectively.

\section*{Future Work and Next Steps}
The critical next phase involves completing the planned experiments:
\begin{enumerate}
    \item \textbf{Universal SynLexLM Model Training:} Fine-tune the Gemma using the combined real and synthetic data multiple datasets under the curriculum learning strategy to see performance of general legal LLM across task types.
    \item \textbf{Rigorous Evaluation:} Evaluate the trained SynLexLM model quantitatively on downstream tasks using benchmarks like BigLaw-Bench \cite{HarveyBigLaw}, and potentially CUAD \cite{CUAD} (QA, classification, contract review) using metrics such as F1, ROUGE \cite{ROUGE}, and accuracy. Compare performance and efficiency against baselines to show the generalizability of our training process.
    \item \textbf{Expand Fine-tuning Data:} Explore fine-tuning on a broader set of legal data, including datasets mentioned previously (US law, German NER \cite{GermanNER}, Australian cases \cite{KaggleLegal}) to assess robustness and adaptability.
    \item \textbf{Refine Methods:} Investigate improvements in synthetic generation (e.g. generating different types of questions or entire legal texts) and curriculum design (e.g. more adaptive complexity metrics).
    \item \textbf{Ethical Considerations:} Continue to analyze and mitigate potential biases in data and model output, and explore explainability techniques for transparency.
\end{enumerate}
Successfully demonstrating the effectiveness of SynLexLM through these evaluations would validate our approach and contribute valuable techniques for developing specialized AI models in data-constrained, high-expertise domains.

\section*{Conclusion}
\label{sec:conclusion}

This project introduced SynLexLM, a strategy aimed at overcoming the data scarcity challenge in training Large Language Models for the complex legal domain. By combining curriculum learning with targeted synthetic data augmentation using large generative models (Gemini Pro), we propose a data-efficient pathway to develop capable legal LLMs.

Our preliminary work successfully established a pipeline for generating diverse synthetic Question-Answer pairs from real legal documents (EurLex, EurLex-Sum) and outlined a clear experimental plan for evaluation against baselines. Qualitative analysis of the generated data indicates its potential relevance for training legal reasoning.

\section*{Team Contributions}

All team members contributed to the project's conceptualization, discussions, and report writing. Regular meetings were held to coordinate efforts, troubleshoot issues, and refine the project direction collaboratively.

 \newpage 

{
\bibliography{references} 

}

\clearpage
\appendix
\section*{Appendix: Sample Synthetic Questions}

\subsection*{EurLex-Sum Long Context Question}
\begin{strip}
    \centering 
    \captionof{table}{Sample Synthetic Question for Text ID eurlex\_sum\_283} 
    \label{tab:eurlex_sum_283_details} 
    \begin{tabular}{@{}lp{0.8\textwidth}@{}}
        \toprule
        \midrule
        \scriptsize \texttt{question} & \scriptsize What organization is involved in this legal matter? \\
        \midrule
        \scriptsize \texttt{keyword} & \tiny 5.12.2009\; EN\; Official Journal of the European Union\; C 296/4\; COUNCIL RECOMMENDATION\; of 30 November 2009\; on smoke-free environments\; 2009/C 296/02\; THE COUNCIL OF THE EUROPEAN UNION,\; Having regard to the Treaty establishing the European Community, and in particular the second subparagraph of Article 152(4) thereof,\; Having regard to the proposal from the Commission,\; After consulting the European Parliament (1),\; Having regard to the opinion of the European Economic and Social Committee (2),\; Whereas:\; (1) Article 152 of the Treaty stipulates that Community action, which shall complement national policies, shall be directed towards improving public health, preventing human illness and diseases, and obviating sources of danger to human health.\; (2) According to Article 137 of the Treaty, the Community shall support and complement the activities of the Member States, inter alia, in the field of improvement in particular of the working environment to protect workers’ health and safety.\; (3) Exposure to environmental tobacco smoke (ETS) — also referred to as second-hand tobacco smoke — is a widespread source of mortality, morbidity and disability in the European Union.\; (4) According to conservative estimates, 7 300 adults including 2 800 non-smokers died as a result of ETS exposure at their workplace in the European Union in 2002. A further 72 000 adult deaths, including those of 16 400 non-smokers, were linked to ETS exposure at home.\; (5) Exposure to second-hand tobacco smoke is particularly dangerous to children and adolescents and could increase the likelihood of their taking up smoking.\; (6) ETS has been classified as a known human carcinogen by the World Health Organization (WHO) International Agency for Research on Cancer and as an occupational carcinogen by Finland and Germany.\; (7) All people have the right to a high level of health protection and should be protected from exposure to tobacco smoke.\; (8) Voluntary policies at national level have proved ineffective in reducing exposure to tobacco smoke. Member States’ binding legislation, properly enforced and monitored is an effective means of adequately protecting people from the health risks of second-hand tobacco smoke.\; (9) Legislation on smoke-free environments is most effective when it is backed up by measures such as awareness-raising campaigns, support for cessation of tobacco use, strong health warnings on tobacco product packaging and other regulation on tobacco products.\; (10) Civil society has an important role in building support for and ensuring compliance with legislation on smoke-free environments.\; (11) Smoke-free policies should have adequate instruments to implement the multi-sectorial approach to tobacco control.\; (12) There is a need for strengthened cooperation between Member States to facilitate the exchange of information and best practice and develop a standardised EU monitoring system.\; (13) The resolution of the Council and the Ministers for Health of the Member States, meeting within the Council of 18 July 1989 on banning smoking in places open to the public (3) invited the Member States to take measures banning smoking in certain enclosed premises open to the public, and to extend the ban on smoking to all forms of public transport.\; (14) Council Recommendation 2003/54/EC of 2 December 2002 on the prevention of smoking and on initiatives to improve tobacco control (4) recommended that Member States implement legislation and/or other effective measures that provide protection from exposure to ETS in indoor workplaces, enclosed public places, and public transport.\; (15) Council Directive 89/391/EEC of 12 June 1989 on the introduction of measures to encourage improvements in the safety and health of workers at work (5), while not explicitly referring to tobacco smoke, covers all risks to the health and safety of workers (6).\; (16) In its Environment and Health Action Plan (2004-2010) (7), the Commission has undertaken to ‘develop work on improving indoor air quality’, in particular by ‘encouraging the restriction of smoking in all workplaces by exploring both legal mechanisms and health promotion initiatives at both European and Member State level’.\; (17) The consultation initiated by the Commission's Green Paper ‘Towards a Europe free from tobacco smoke: policy options at EU level’ (8) (the ‘Green Paper’) has revealed strong support both for comprehensive smoke-free policies in all enclosed workplaces and public places and for further EU action to promote smoke-free environments throughout the Member States.\; (18) The Employment, Social Policy, Health and Consumer Affairs Council held an exchange of views on policy options at EU level on tobacco smoke-free environments on 30 and 31 May 2007. It welcomed the Green Paper and stressed the need for Community guidance to further promote tobacco-smoke free environments at EU level, as well as Community support for and coordination of national measures.\; (19) The European Parliament's resolution of 24 October 2007 on the Green Paper called on the Member States to introduce comprehensive smoke-free laws within two years and invited the Commission to table a relevant legislative proposal by 2011 in the event of unsatisfactory progress. It also called on the Commission to propose an amendment to the current legislative framework in order to classify ETS as a carcinogen and oblige employers to ensure that the workplace is smoke-free.\; (20) Article 8 of the WHO Framework Convention on Tobacco Control (FCTC), signed in June 2003 by all WHO members, and so far ratified by 167 Parties, including the Community and 26 of its Member States, creates a legal obligation for its Parties to adopt and implement in areas of existing national jurisdiction as determined by national law and to actively promote, at other jurisdictional levels, the adoption and implementation of effective measures to protect people from exposure to second-hand tobacco smoke in all indoor workplaces, public transport and indoor public places and, as appropriate, other public places.\; (21) The Second Conference of the Parties to FCTC in July 2007 adopted guidelines on protection from exposure to tobacco smoke (9) to assist Parties in meeting their obligations under Article 8 of the Convention. Each Party should strive to implement the guidelines within five years of the Convention's entry into force for that Party.\; (22) Article 14 of the WHO Framework Convention creates a legal obligation for its Parties to develop and disseminate appropriate, comprehensive and integrated guidelines based on scientific evidence and best practices and to take effective measures to promote the cessation of tobacco use and adequate treatment for tobacco dependence. The Third Conference of the Parties to the WHO Framework Convention decided to establish a working group for the elaboration of guidelines for implementation of that Article.\; (23) The European Strategy on Tobacco Control adopted by the WHO Regional Committee for Europe in September 2002 recommended that Member States ensure the citizens’ right to a smoke-free environment by, inter alia, making public places, workplaces and public transport smoke-free, banning smoking outdoors in all educational institutions for minors, in all places of healthcare delivery and at public events, as well as classifying ETS as a carcinogen.\; (24) This Recommendation is without prejudice to the Community legislation laying down minimum requirements for the safety and health protection of workers adopted under Article 137 of the Treaty, to Directive 2001/37/EC of the European Parliament and of the Council of 5 June 2001 on the approximation of the laws, regulations and administrative provisions of the Member States concerning the manufacture, presentation and sale of tobacco products (10) and to Commission Decision 2003/641/EC of 5 September 2003 on the use of colour photographs or other illustrations as health warnings on tobacco packages (11),\; HEREBY RECOMMENDS THAT THE MEMBER STATES:\; 1. provide effective protection from exposure to tobacco smoke in indoor workplaces, indoor public places, public transport and, as appropriate, other public places as stipulated by Article 8 of the WHO Framework Convention on Tobacco Control (FCTC) and based on the annexed guidelines on protection from exposure to tobacco smoke adopted by the Second Conference of the Parties to FCTC, within five years of the FCTC’s entry into force for that Member State, or at the latest within three years following the adoption of this Recommendation;\; 2. develop and/or strengthen strategies and measures to reduce exposure to second-hand tobacco smoke of children and adolescents;\; 3. complement smoke-free policies with supporting measures, which may include:\; (a) taking effective measures to promote cessation of tobacco use and adequate treatment for tobacco dependence, taking into account national circumstances and priorities as outlined in Article 14 of the FCTC; and\; (b) introducing combined warnings, as defined by Article 2(4) of Commission Decision 2003/641/EC of 5 September 2003 on the use of colour photographs or other illustrations as health warnings on tobacco packages (12), and information on services supporting the cessation of tobacco use on the packages of smoking tobacco products in order to better inform consumers about the health risks of tobacco use and exposure to tobacco smoke, encourage cessation of tobacco use and deter initiation;\; 4. develop, implement, periodically update and review comprehensive multi-sectoral tobacco control strategies, plans or programmes which address, inter alia, the issue of protection from tobacco smoke in all places accessible to the general public or places of collective use, regardless of ownership or right to access;\\
        \midrule
        \scriptsize \texttt{answer} & \scriptsize Based on the provided text, several organizations are involved in this legal matter, directly or indirectly: 
            \begin{itemize}[itemsep=0cm]
                \item Council of the European Union: This is the primary organization issuing the recommendation.
                \item European Commission: They proposed the recommendation.
                \item European Parliament: Consulted on the recommendation.
                \item European Economic and Social Committee: Provided an opinion on the recommendation.
                \item World Health Organization (WHO): The WHO Framework Convention on Tobacco Control (FCTC) is referenced and plays a significant role in the recommendation.
                \item WHO Regional Committee for Europe: Their European Strategy on Tobacco Control is mentioned.
                \item Member States of the European Union: The recommendation is addressed to them, and they are expected to implement the measures.
            \end{itemize}
            Therefore, the main organization involved in issuing the legal document is the \textbf{Council of the European Union}. \\
        \midrule
        \scriptsize \texttt{type} & \scriptsize factual \\
        \midrule
        \scriptsize \texttt{metadata.celex\_id} & \scriptsize 32009H1205(01) \\ 
        \midrule
        \scriptsize \texttt{difficulty} & \scriptsize easy \\
        \bottomrule
    \end{tabular}
\end{strip}

\clearpage
\subsection*{EurLex Reasoning Question}
\begin{strip}
    \centering 
    \captionof{table}{Sample Synthetic Question for Text ID eurlex\_15107} 
    \label{tab:eurlex_15107_details} 
    \begin{tabular}{@{}lp{0.8\textwidth}@{}}
        \toprule
        \midrule
        \scriptsize \texttt{question} & \scriptsize How does the document justify the decision concerning this? \\
        \midrule
        \scriptsize \texttt{keyword} & \scriptsize Whereas pending the entry into force of the Economic Partnership, Political Coordination and Cooperation Agreement signed in Brussels on 8 December 1997, the Interim Agreement on trade and trade-related matters between the European Community, of the one part, and the United Mexican States, of the other part, signed in Brussels on 8 December 1997, should be approved, \\
        \midrule
        \scriptsize \texttt{answer} & \scriptsize The document justifies the decision to approve the Interim Agreement on trade and trade-related matters by stating that it is necessary \textbf{pending the entry into force of the broader Economic Partnership, Political Coordination and Cooperation Agreement.} \newline\newline In simpler terms, the Interim Agreement is seen as a temporary measure to establish trade relations while the more comprehensive and long-term agreement is being finalized and put into effect. It bridges the gap and allows trade to commence before the larger agreement is fully implemented. \\ 
        \midrule
        \scriptsize \texttt{type} & \scriptsize reasoning \\
        \midrule
        \scriptsize \texttt{metadata.labels} & \scriptsize 9, 23, 93, 96, 97 \\ 
        \midrule
        \scriptsize \texttt{difficulty} & \scriptsize hard \\
        \bottomrule
    \end{tabular}
\end{strip}
\end{document}